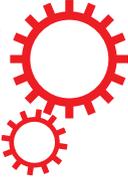

# OPEN  Computer-aided implant design for the restoration of cranial defects

Xiaojun Chen[1], Lu Xu[1], Xing Li[1] & Jan Egger[2,3]

Patient-specific cranial implants are important and necessary in the surgery of cranial defect restoration. However, traditional methods of manual design of cranial implants are complicated and time-consuming. Our purpose is to develop a novel software named EasyCrania to design the cranial implants conveniently and efficiently. The process can be divided into five steps, which are mirroring model, clipping surface, surface fitting, the generation of the initial implant and the generation of the final implant. The main concept of our method is to use the geometry information of the mirrored model as the base to generate the final implant. The comparative studies demonstrated that the EasyCrania can improve the efficiency of cranial implant design significantly. And, the intra- and inter-rater reliability of the software were stable, which were 87.07 ± 1.6% and 87.73 ± 1.4% respectively.

The cranial defect restoration is commonly performed when a patient suffers from head trauma due to an accident or injury[1–3]. The main aims of cranial defect restoration are to protect of the brain or improve the cranial appearance. RP (Rapid Prototype) skull models based on 3D CT data have been used to manually fabricate implants for decades[4]. However, many disadvantages are shown by using the implants, including expensive cost, massive time, and complicated procedures[5]. CAD/CAM techniques has been introduced to prefabricate tailored implants for decades which are precisely formed based on the CT data of the patient[6, 7]. Most Surgeons agree that the techniques provide precise planning of the cranial defect restoration surgery, and the implants are reliable with high grades biocompatibility and mechanical properties[8]. However, the traditional CAD method using general softwares such as UG (Siemens PLM Software, Plano, USA), SolidWorks (Dassault Systèmes SOLIDWORKS Corp., Waltham, USA), Magic RP and 3-Matic (Materialise, Leuven, Belgium), Geomagic (3D SYSTEMS, Rock Hill, USA), etc.[9, 10] is time consuming and complicated.

For example, Jardini *et al.*[10] presented a clinical application of the design and fabrication of a biomodel and customized implant for the surgical reconstruction of a large cranial defect using some commercial softwares such as Mimics, Magics and SolidWorks. Firstly, the CT image data were processed and converted through Mimics, and a 3D cranial model was reconstructed subsequently. Then, the 3D model was edited using Magics 15.0 software in order to minimize surface imperfections, and Boolean operations were then applied to reconstruct the cranial defect surface. After that, the new generated point cloud that defined the implant 3D geometry was processed to develop a CAD surface model. Finally, on the basis of this surface model, the customized implant for cranial reconstruction was designed in the SolidWorks software.

As for the 3-Matic software, the functions of combining surfaces, repairing and de-featuring, remeshing, modifying and editing, etc. are used for the implant design. Although the complexity level of the usage of the 3-Matic is lower than above mentioned method (for example, no need of importing and exporting), the user is still required to own the engineering background knowledge of geometry design and get very familiar with it. For this reason, it may be still too complicated and difficult for a surgeon to learn.

In addition, Chulvi *et al.*[11] proposed a system for automatically design the customized implants by linking two computer prototypes. The research used the Knowledge Based System technology to store and manage medical data and combined many existing software, including Mimics, 3D Slicer, ImLib3D, MITK, OsiriX and VTK. However, there is no further description about the detail algorithm. Scharver *et al.*[12] introduced a prototype system based on augmented reality to design the cranial implant. Yet several steps remain labor-intensive and expensive.

In this study, a novel software named EasyCrania for the design of cranial implant has been developed based on some well-known open-source toolkits including VTK (http://www.vtk.org/) and Qt (http://qt-project.org/).

[1]Institute of Biomedical Manufacturing and Life Quality Engineering, State Key Laboratory of Mechanical System and Vibration, School of Mechanical Engineering, Shanghai Jiao Tong University, Shanghai, China. [2]Faculty of Computer Science and Biomedical Engineering, Institute for Computer Graphics and Vision, Graz University of Technology, Graz, Austria. [3]BioTechMed-Graz, Graz, Austria. Correspondence and requests for materials should be addressed to X.C. (email: xiaojunchen@163.com) or J.E. (email: egger@icg.tugraz.at)

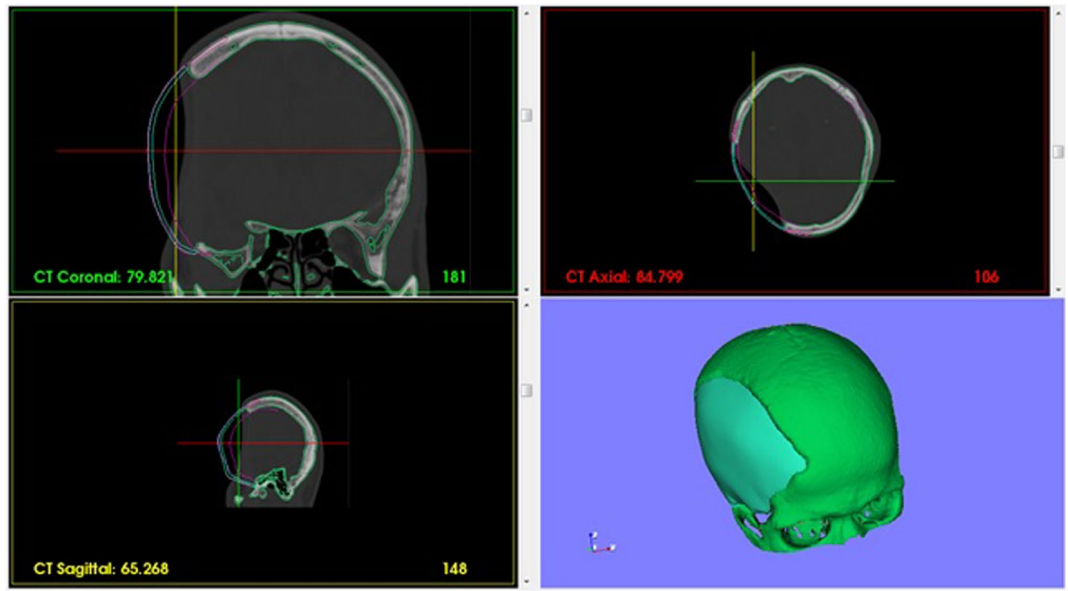

**Figure 1.** Evaluating one case of the cranial implant with the CT data of the patient.

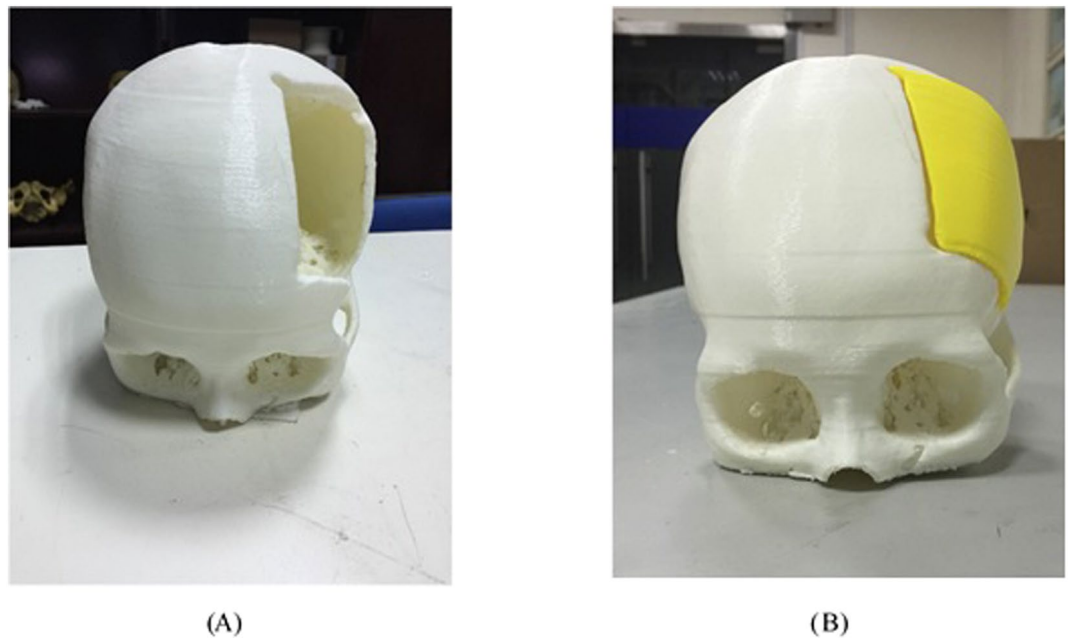

**Figure 2.** The 3D printed crania model and implant.

By adopting easier and more efficient approach, the time and complexity of designing the implant has been massively reduced. The comparison with traditional design method has been conducted and demonstrated the efficiency of our method.

## Results

A general framework of the cranial implant design was introduced and several algorithms were presented, including mirroring model, clipping and surface fitting. A software named EasyCrania was developed under the platform of Microsoft Visual Studio 2010 (Microsoft, Washington, USA). Some famous open source toolkits including VTK (Visualization Toolkit, an open-source library for 3D computer graphics, image processing and visualization, http://www.vtk.org/) and Qt (a cross-platform application and UI framework, http://www.qt-project.org/) were involved. Several cases of customized cranial implant design were conducted using EasyCrania. Two cases are shown in Figs 1 and 2.

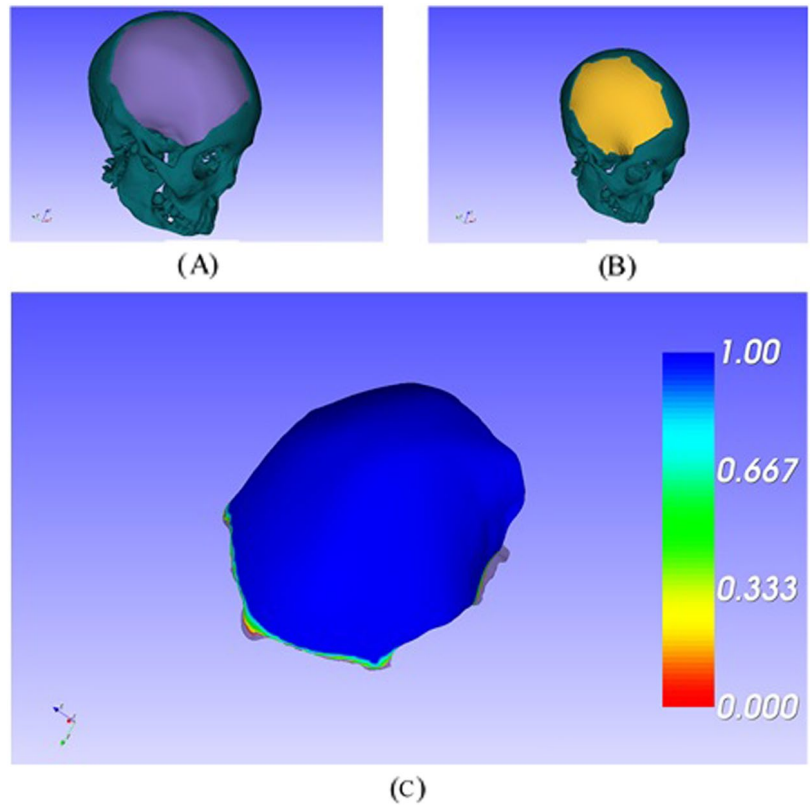

**Figure 3.** The comparison between the implants designed using EasyCrania and traditional softwares: (**A**) the implant designed using EasyCrania, (**B**) the implant designed using traditional softwares, (**C**) the distance filed between the two implants.

In addition, twenty participants designed the implant for the same cranial defect case using traditional softwares (including Mimics, Imageware, UG together) and EasyCrania respectively (shown in Fig. 3). The workflow of the former method is described as follows:

1. On the basis of the original CT data, image segmentation is performed in Mimics software, and a 3D-reconstructed model of the crania can be obtained and exported in the format of ".stl".
2. The 3D-reconstructed cranial model is then imported to the Imageware software. The user needs to manually draw the curves through the complicated functions of "creating and modifying new curves and surfaces", aiming at extracting the data of the cranial defect for surface fitting. In order to achieve the high quality result of surface fitting, it is also required to process the initial curved surface such as the removal of surface imperfections and smoothing treatment. Since the surface consists of triangle meshes, the function of "Auto surface" is used to optimize it and the result was exported as file in '.imw' format.
3. Based on the '.imw' file of above procedure, the solid model of the implant was created through the functions of "creating models though curves and polylines" in the UG software, and the result was then exported as file in '.stl' format for the manufacture.

With respect to the later method, the user only needs to load the CT data to the EasyCrania firstly and the 3D-reconstructed model of the crania can be achieved through the image processing procedure. Secondly, the reference model can be obtained through a specified mirroring plane and the part of surface over the defect of the cranial model can be extracted. Then, some initial control points are indicated and the contour curve can be generated and updated dynamically, so that the clipped model can be extracted through contour clipping (also known as the surface fitting procedure). Finally, the initial implant with a smoother inner surface and the final implant fitting the edge of the defect are generated automatically.

The consumed time for each implant design through these two methods was measured respectively and the mean value was calculated. According to the results of statistical analysis and evaluation, the mean time of using several traditional softwares was $137.8 \pm 5$ mins, while the mean time of using EasyCrania was only $26.7 \pm 2$ mins. The distance field of the two designed implants through two methods indicated that they have no significant difference near the margin of the defect, which demonstrated that the implant designed by EasyCrania can fit the cranial model well. Furthermore, as the implant was generated base on the mirrored model, it was more symmetric and aesthetic than the one designed by traditional softwares.

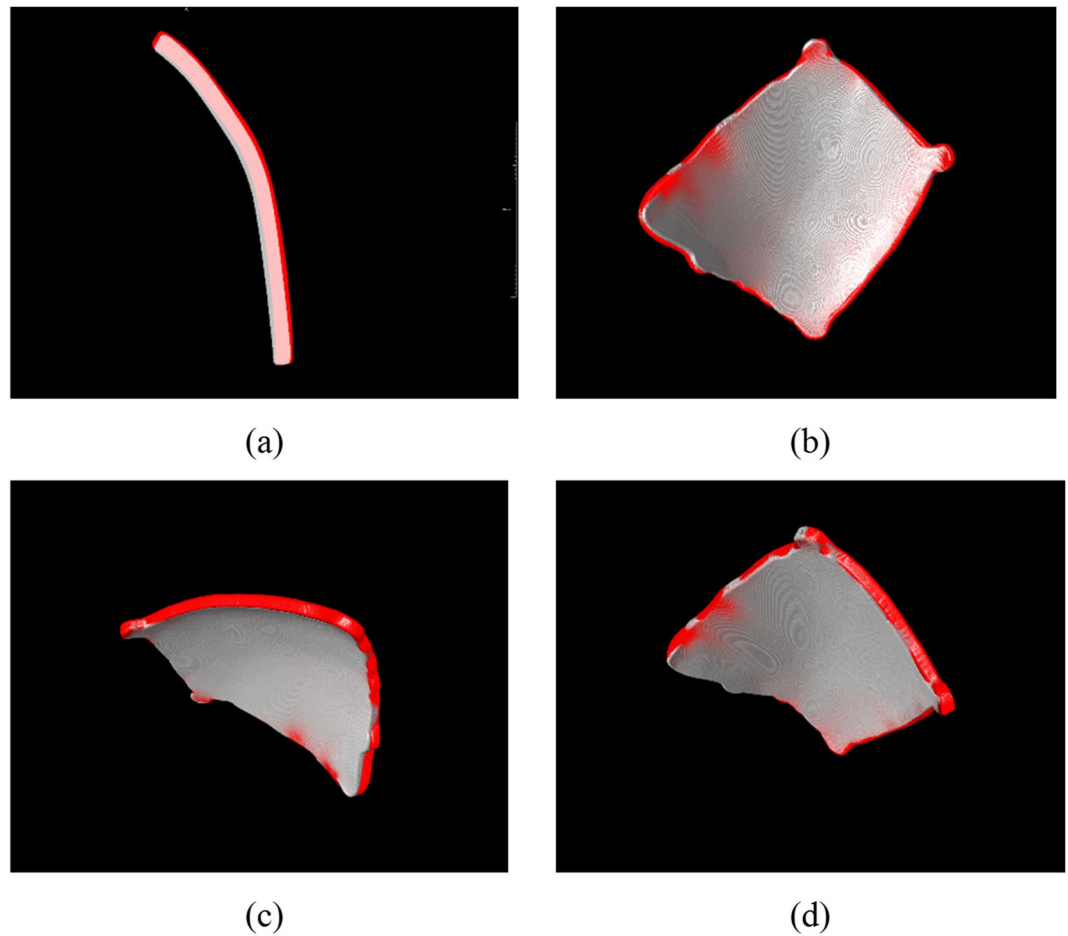

**Figure 4.** The calculation of overlapping rate.

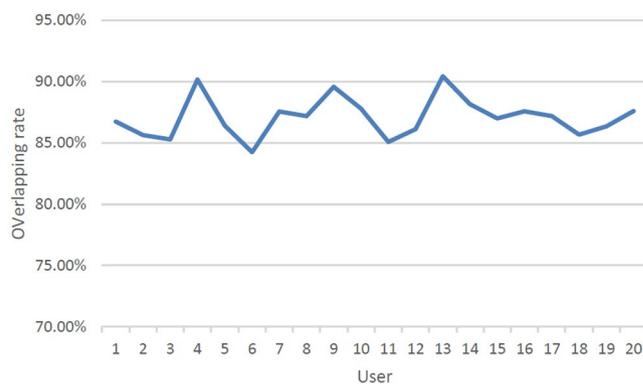

**Figure 5.** The overlapping rate of twenty users.

In terms of the intra-rater reliability, the results were imported to a specific self-developed software using VTK for calculating the overlapping rate (shown in Fig. 4). The results demonstrated that the intra-rater reliability was stable (shown in Fig. 5), and the mean value was 87.07 ± 1.6%. With respect to the inter-rater reliability, one user have conducted the implant design for twenty times in the same case, and the mean inter-rater reliability was 87.73 ± 1.4% (shown in Fig. 6).

### Discussion and Conclusion

Over the past years, the titanium mesh has been used as intraoperative malleable substitutes for the improvement of neurologic function in cranioplasty. However, it is expensive, and time-consuming due to the intraoperative mixing for the preparation, adaptation, and contouring of the implant for the defect[13]. Recently, with the development of medical imaging and 3D printing technology, the polyetheretherketone (PEEK) customized implant has been manufactured to improve surgical outcome. Kim et al. (2012) prefabricated customized cranioplasty implants for 16 patients with large skull defects using MIMICS 13.1 Software (Materialise Inc., Leuven, Belgium) and Spectrum Z 510 3D Printer (Z Corporation, Burlington, MA, USA)[13]. The results demonstrated that it is a cost-effective technique to overcome the shortcomings of intraoperative molding for the reconstruction of various cranial defects. Furthermore, the median operation time can be reduced significantly compared with the conventional non-customized implants. However, the design procedure of patient-specific implant based on the computer-aided design (CAD) technique is time-consuming and complicated, requiring high level of the engineering background for the users.

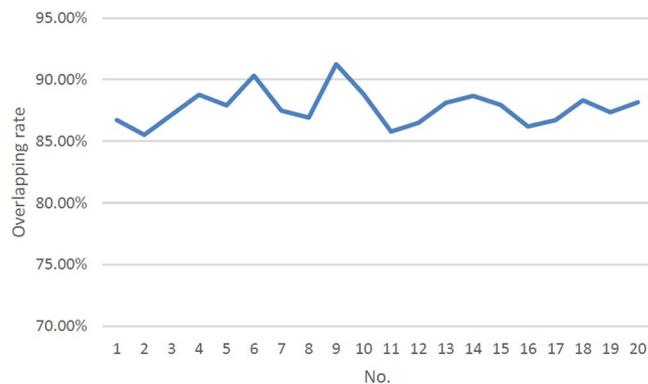

**Figure 6.** The overlapping rate of implant design for twenty times in the same case.

Therefore, a semi-automatic software named EasyCrania was developed in this study focusing on the user-friendliness for the surgeons. The EasyCrania can provide an easy and efficient approach to design a customized implant for cranial defect restoration surgery based on the 3D cranial model of the patient. The mirrored model can ensure the shape of the implant to be aesthetic looking, and the specified contour can guarantee the implant fits the edge of the defect well. Only three manual steps are needed, including choosing the anatomically symmetric points to generate the median plane and indicating two contours, which is much easier and simpler than the traditional ways. In conclusion, the achieved highlights of this study are:

- Efficient planning and reconstruction of the implant for cranial defect restoration;
- Mirroring skull based on the anatomically symmetric points of the model;
- Triangulation for an aesthetic looking initial implant;
- Evaluation the implant with the real CT data of the patient for clinical use;
- Exporting the implant as STL file for 3D printing.

Compared with the commercially available softwares, there is a trend of using some well-known open-source toolkits (such as VTK and Qt) in the development of computer-assisted surgery softwares. As VTK provides various kinds of new emerging, leading-edge algorithms in the field of medical image computing such as GPU-accelerated volume rendering, automatic functionality segmentation using the level set method, efficient non-rigid registration for multi-modality imaging, etc., the EasyCrania can be easily extended to meet individual complicated requirements of the surgeons[14].

Currently, the EasyCrania is just in a preliminary stage and need clinical trials to evaluate the reliability. In addition, as Delaunay triangulation algorithm is used to accomplish the surface fitting, if the surface is not convex, the triangulated surface may fail to fit the cranial defect. Nevertheless, future work still need to be done to improve the robustness of EasyCrania. Also, the prototype we proposed can be further developed with MeVisLab[15, 16]. As our research is based on some open source toolkits, it will be added into the extension module of 3D Slicer (an open-sourced software platform for visualization and medical image computing, http://www.slicer.org/) to be shared by the global community. Furthermore, since the VTK and Qt support for different operating systems (Windows, Linux, and Mac Os X), our software will be easily developed to a multi-platform application environment[17]. In addition, on the basis of these algorithms, our novel technique may also be extended to other kinds of surgeries such as the reconstruction of maxillofacial defects in the future.

## Methods

The CT data is firstly loaded in EasyCrania and the 3D model of the crania can be achieved by processing image threshold segmentation, region growing and Flying Edges[18]. Based on the 3D cranial model, the mirrored model can be realized along the middle plane defined by anatomically symmetric points. Then, a surface is fitted on the basis of the region surrounding the skull defect, and the final implant is generated using surface clipping, surface offsetting, and ruled surface construction. The framework of EasyCrania is shown in Fig. 7.

**Mirroring Model.** This section mainly discusses how to achieve the mirrored model on the base of the 3D cranial model. Firstly, several anatomically symmetric points of the skull model are selected to generate the median plane. Obviously, we need to calculate the transform matrix T of the mirroring. Instead of directly calculate the matrix, we break the process into several steps. We firstly transform the plane origin to the coordinate origin, suppose the matrix of this step as T1, then transform the normal of the median plane along the X axis, and suppose the matrix is T2. The next step is to calculate the matrix T3 of rotating 180° around the Y axis. Then by inversing the process, we can acquire the matrix T, shown in Equation 1. One case of mirroring models is shown in Fig. 8.

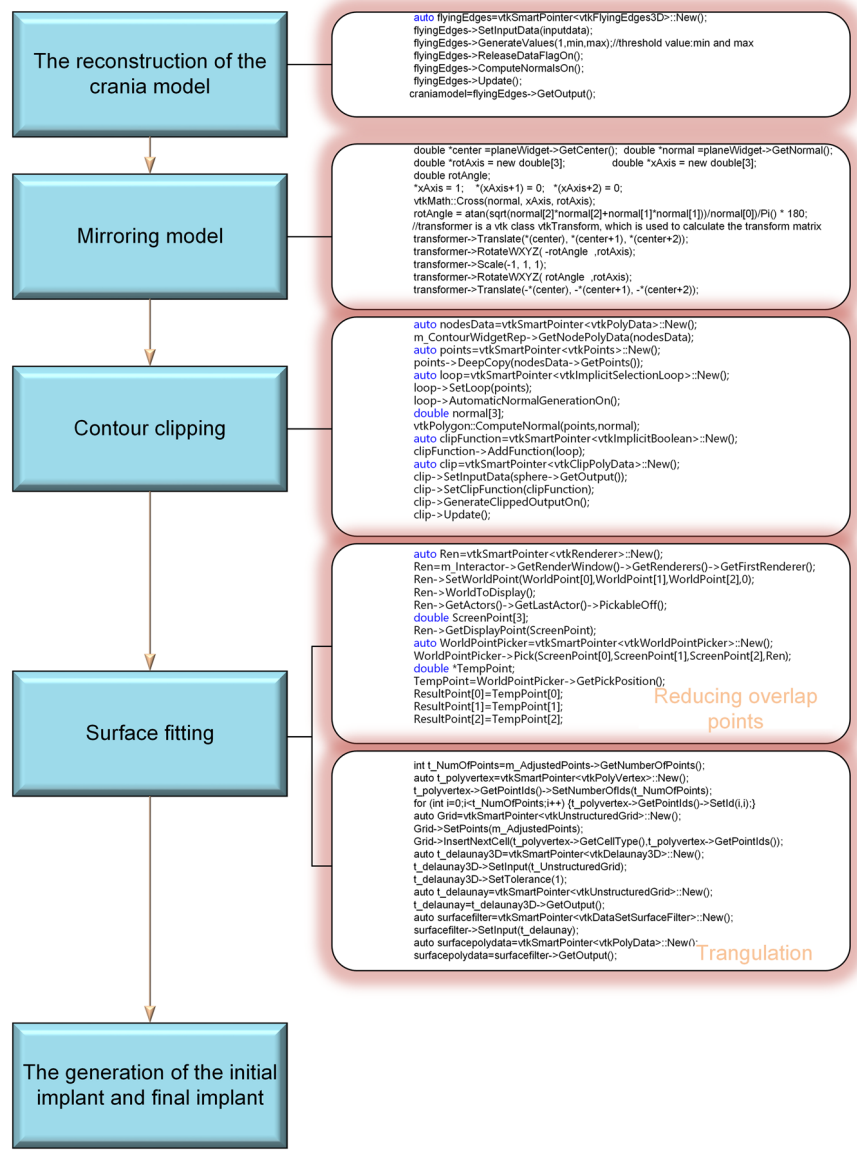

**Figure 7.** The architecture of EasyCrania.

$$T = T_1 T_2 T_3 T_2' T_1' \qquad (1)$$

**Contour clipping.** The clipping is conducted to extract the part of surface over the defect of the cranial model from the mirrored model. Firstly, a non-self-intersecting contour is indicated along the edge of the defect (Shown in Fig. 9(A)), then an implicit function is computed based on the contour (Shown in Fig. 9(B)). The implicit function is defined by the direction vector and the area of the contour. The direction vector is computed based on the normal of all the nodes on the contour. The clipping is accomplished by using a VTK class vtkClipPolydata, which takes an implicit function as the clipping function. The class would calculate the signed distance field of the model for clipping. If the cell of the model for clipping is inside the implicit function, the scalar value of it would be negative. If the cell is outside the implicit function, the scalar value would be positive. If the cell is on the implicit function, then the scalar value is set to be zero. Then based on the scalar value, the model inside the implicit function can be clipped out (Shown in Fig. 9(C,D)).

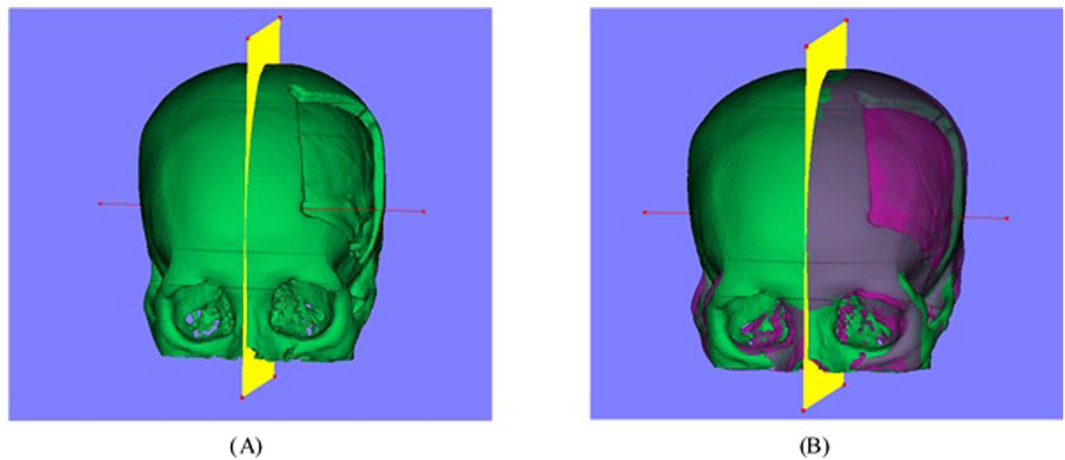

**Figure 8.** Mirroring model: (**A**) the generated median plane, (**B**) the mirrored model based on the median plan and the cranial model.

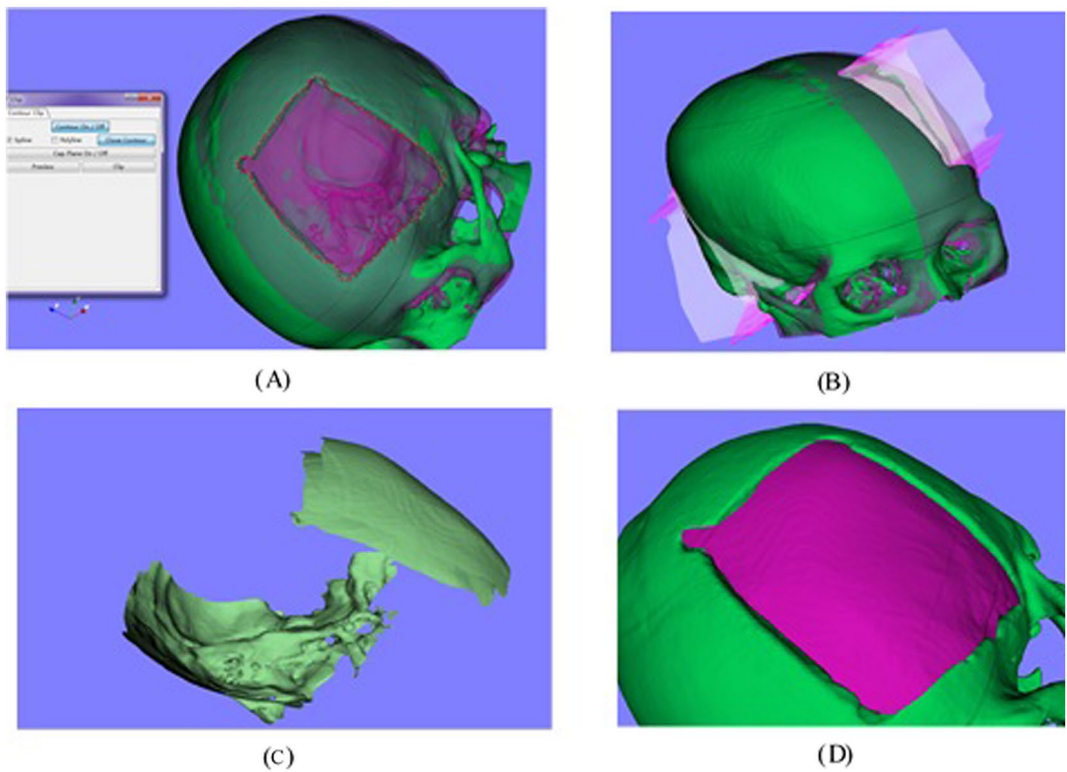

**Figure 9.** Contour clipping: (**A**) the contour along the defect edge, (**B**) the implicit function computed from the contour, (**C**) the clipped model and (**D**) the view of the crania model and the clipped model.

**Surface fitting.** This section discusses how to reconstruct a surface combined the geometry information of the cranial model and the clipped model extracted through contour clipping. Delaunay triangulation algorithm[19] is used to fit the surface based on the points, which were selected by the indicated contour. The contour would specify the area for triangulation. In order to reduce the overlap points, all the points within the contour are firstly projected on the screen, and then the view coordinates are converted into world coordinates, where all the points are located on the surface at the present view angle. One case of surface fitting is shown in Fig. 10(A,B), and the algorithm pseudocode is described as follows:

Define SPList as the list of the points from the surface, and N as the number of the points.

Define DPList as the display coordinates of the points from the surface, and M as the number of the display points.

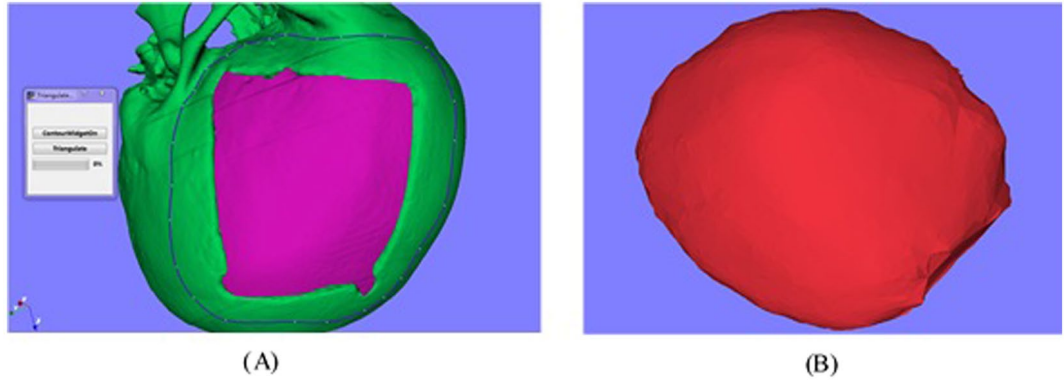

**Figure 10.** Surface fitting: (**A**) the contour indicate the area for Delaunay triangulation, (**B**) the triangulated surface.

```
Define FinalList as the points for surface fitting.
Define Surfacepolydata is the result of surface fitting.
for i to N
{
    P_i is the i^th point of SPList;
    D_i is the i^th display coordinate point;
    vtkRenderer *Ren;
    Ren->SetWorldPoint(P_i);
    Ren->WorldToDisplay();
    Ren->GetDisplayPoint(D_i);
    if (!DPList.iscontain(D_i))
        {
            DPList.append(D_i);
        }
}
for i to M
{
    P_i is the i^th point of FinalList;
    D_i is the i^th point of DPList;
    vtkWorldPointPicker *WorldPointPicker;
    WorldPointPicker->Pick(D_i,Ren);
    P_i = WorldPointPicker->GetPickPosition();
    FinalList.append(P_i);
}
vtkPolyVertex *t_polyvertex;
t_polyvertex->GetPointIds()->SetNumberOfIds(M);
for i to M
    {
        t_polyvertex->GetPointIds()->SetId(i,i);
    }
vtkUnstructuredGrid *Grid;
Grid->SetPoints(FinalList);
Grid->InsertNextCell(t_polyvertex->GetCellType(),t_polyvertex->GetPointIds());
vtkDelaunay3D *t_delaunay3D;
t_delaunay3D->SetInput(t_UnstructuredGrid);
t_delaunay3D->SetTolerance(1);
vtkUnstructuredGrid *t_delaunay;
t_delaunay = t_delaunay3D->GetOutput();
vtkDataSetSurfaceFilter *surfacefilter;
surfacefilter->SetInput(t_delaunay);
vtkPolyData *Surfacepolydata;
Surfacepolydata = surfacefilter->GetOutput();
```

**The generation of the initial implant.** The outer surface of the reconstructed model through Delaunay triangulation is rough, and the inner surface is poor. However, the shape of the outer surface fit the cranial model well. Chen et al.[20] introduced a semi-automatic method for template design, mainly consists tree steps including inner surface generation, outer surface generation and ruled surface generation[20]. The input of the method is a model as the base of generating the template and a contour which indicates the shape of the template. In order to generate the initial implant, the contour indicated in surface fitting and the triangulated model are used as the input of the method. The generated initial implant is shown in Fig. 11.

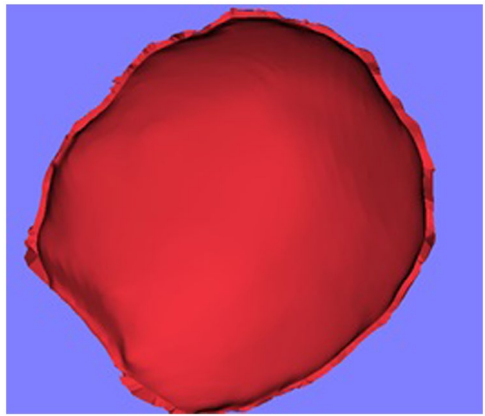

**Figure 11.** The generated initial implant.

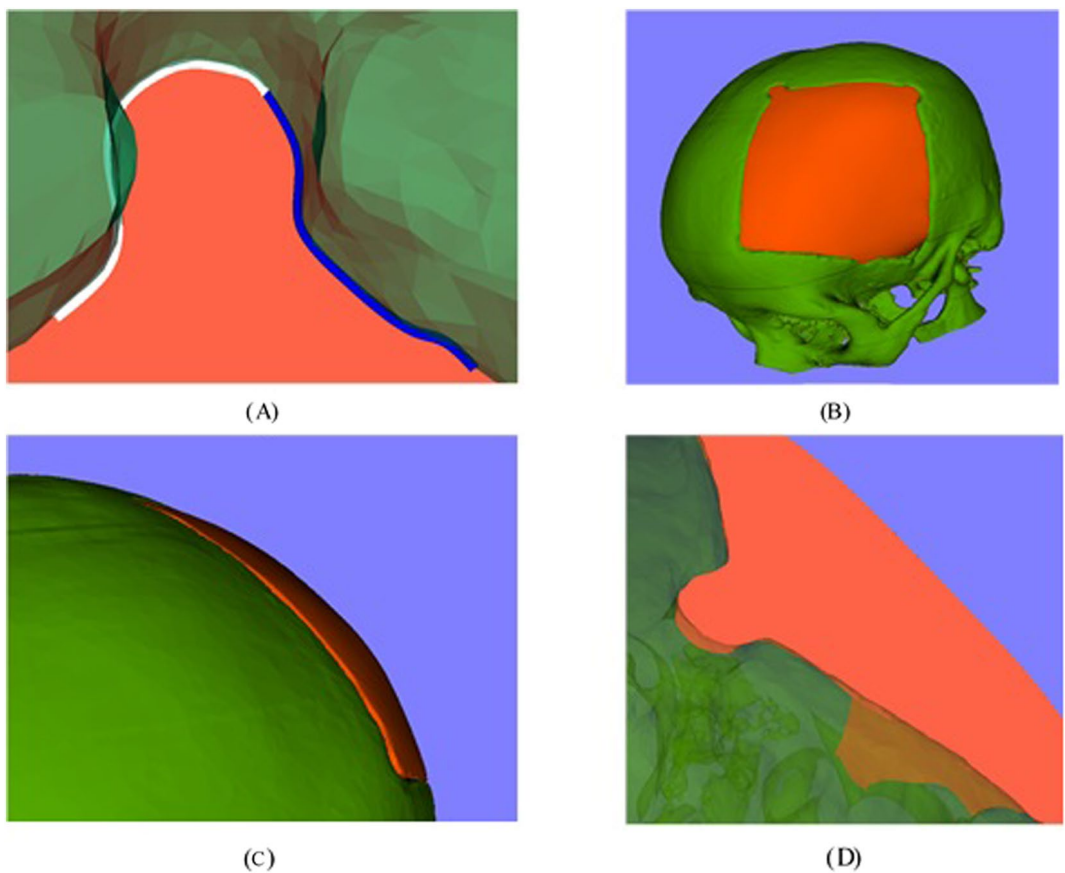

**Figure 12.** The final implant: (**A**) the contour along the edge of the defect from inner surface of the crania model, (**B**) the final implant and the crania model, (**C**) and (**D**) different view of the final implant.

**The generation of the final implant.** The initial implant has a smoother inner surface, but the shape doesn't fit the edge of the defect. By adopting the same method in the generation of the initial implant, the final implant with a smooth inner-outer surface can be achieved. In order to fit the cranial model at the edge, a contour is indicated based on the shape of defect from inside of the cranial model. The generated final implant is shown in Fig. 12.

### References

 1. Dujovny, M., Aviles, A., Agner, C., Fernandez, P. & Charbel, F. T. Cranioplasty: Cosmetic or therapeutic? *Surgical Neurology* **47**(3), 238–41 (1997).
 2. Pompili, A. *et al.* Cranioplasty performed with a new osteoconductive osteoinducing hydroxyapatite-derived material. *J. Neurosurg.* **89**, 236–42 (1998).
 3. Mohamed, N., Majid, A. A., Piah, A. R. M., Rajion, Z. A. Designing of skull defect implants using C1 rational cubic Bezier and offset curves. *International Conference On Mathematics, Engineering And Industrial Applications 2014 (ICoMEIA 2014)*. AIP Publishing, **1660**, 050003 (2015).
 4. Min, K. Dean, D. Highly Accurate CAD Tools for Cranial Implants. Medical Image Computing & Computer-Assisted Intervention-Miccai, International Conference, Montréal, Canada, November. 99-107 (2003).
 5. Voigt, M., Schaefer, D. J. & Andree, C. Three-dimensional reconstruction of a defect of the frontozygomatic area by custom made Proplast II implant. *European Journal of Plastic Surgery* **23**(7), 391–394 (2000).
 6. Dean, D., Min, K. J. & Bond, A. Computer aided design of large-format prefabricated cranial plates. *Journal of Craniofacial Surgery* **14**(6), 819–32 (2003).
 7. Müller, A., Krishnan, K. G., Uhl, E. & Mast, G. The application of rapid prototyping techniques in cranial reconstruction and preoperative planning in neurosurgery. *Journal of Craniofacial Surgery* **14**(6), 899–914 (2003).
 8. Hoffmann, B. & Sepehrnia, A. Taylored implants for alloplastic cranioplasty - clinical and surgical considerations. *Acta. neurochirurgica. Supplement* **93**(93), 127–9 (2005).
 9. Meer, W. J. V. D., Bos, R. R. M., Vissink, A. & Visser, A. Digital planning of cranial implants. *British Journal of Oral & Maxillofacial Surgery* **51**(5), 450–2 (2013).
10. Jardini, A. L. *et al.* Cranial reconstruction: 3D biomodel and custom-built implant created using additive manufacturing. *Journal of Cranio-Maxillo-Facial Surgery* **42**, 1877–1884 (2014).
11. Vicente, C. R., David, C. T., Rosario, V. & Alex, S. V. Automated design of customized implants. *Rev.fac.ing.univ.antioquia* **24**(68), 95–103 (2013).
12. Scharver, C., Evenhouse, R., Johnson, A. & Leigh, J. Pre-surgical Cranial Implant Design using the PARIS™ Prototype. *IEEE Virtual Reality IEEE Computer Society* **5**(1), 199–291 (2004).
13. Kim, B. J. *et al.* Customized cranioplasty implants using three-dimensional printers and polymethyl-methacrylate casting. *J. Korean Neurosurg. Soc.* **52**, 541–546 (2012).
14. Chen, X. *et al.* A surgical navigation system for oral and maxillofacial surgery and its application in the treatment of old zygomatic fractures. *Int. J. Med. Robotics Comput. Assist Surg.* **7**, 42–50 (2011).
15. Egger, J. *et al.* Integration of the OpenIGTLink Network Protocol for image-guided therapy with the medical platform MeVisLab. *International Journal of Medical Robotics + Computer Assisted Surgery Mrcas* **8**(3), 282–90 (2012).
16. Gall, M., Li, X., Chen, X. J., Schmalstieg, D., Egger, J. Computer-Aided Planning of Cranial 3D Implants. *Computer Assisted Radiology and Surgery (CARS)* 2016.
17. Chen, X., Yuan, J., Wang, C., Huang, Y. & Kang, L. Modular preoperative planning software for computer-aided oral implantology and the application of a novel stereolithographic template: a pilot study. *Clinical Implant Dentistry and Related Research* **12**(3), 181–193 (2010).
18. Schroeder, W., Maynard, R., Geveci, B. Flying edges: A high-performance scalable isocontouring algorithm, Large Data Analysis and Visualization (LDAV). *2015 IEEE 5th Symposium on. IEEE*.33–40 (2015).
19. Gopi, M., Krishnan, S. & Silva, C. T. Surface Reconstruction based on Lower Dimensional Localized Delaunay Triangulation. *Computer Graphics Forum Blackwell Publishers Ltd.* **19**(3), 467–478 (2010).
20. Chen, X. J., Xu, L., Yang, Y. & Egger, J. A semi-automatic computer-aided method for surgical template design. *Scientific Reports* **6**, 20280–20298 (2016).

### Acknowledgements

This work was supported by the Foundation of Science and Technology Commission of Shanghai Municipality (14441901002, 15510722200, 16441908400), Shanghai Jiao Tong University Foundation on Medical and Technological Joint Science Research (YG2016ZD01). Dr. Dr. Jan Egger receives funding from Bio TechMed-Graz ("Hardware accelerated intelligent medical imaging").

### Author Contributions

X.C., X.L. and L.X. conceived of the project, and designed the framework of the software. X.C., X.L. and L.X. proposed the new algorithms, developed the software and performed the experiments. X.C., L.X., X.L. and J.E. wrote the paper. All authors discussed the results and commented on the manuscript.

### Additional Information

**Competing Interests:** The authors declare that they have no competing interests.

**Publisher's note:** Springer Nature remains neutral with regard to jurisdictional claims in published maps and institutional affiliations.